\documentclass[lettersize,journal]{IEEEtran}  

\IEEEoverridecommandlockouts                              

\usepackage{amsmath} 

\usepackage{hyperref}
\usepackage{graphicx}
\usepackage{stfloats}
\usepackage{caption}
\usepackage{subcaption}
\usepackage{ctable}
\usepackage{cite}
\usepackage{multirow}
\usepackage{amsfonts}
\usepackage{wrapfig}

\newcolumntype{M}[1]{>{\centering\arraybackslash}m{#1}}
\newcolumntype{N}{@{}m{0pt}@{}}

\DeclareUnicodeCharacter{2212}{-}

\title{\LARGE \bf
Traffic-Twitter Transformer: A Nature Language Processing-joined Framework for Network-wide Traffic Forecasting
}

\author{Meng-Ju Tsai$^{1}$, Zhiyong Cui$^{2}$, Hao (Frank) Yang$^{1}$, Cole Kopca$^{1}$, Sophie Tien$^{3}$, and Yinhai Wang$^{1}$
\thanks{*This work was supported by PacTrans and Dr. and Mrs. Paul Liao Endowed Regental Fellowship}
\thanks{$^{1}$M.J. Tsai, H. Yang, C. Kopca, and Y. Wang are with the Department of Civil and Environmental Engineering, University of Washington, Seattle, WA(e-mail: mjtsai@uw.edu; zhiyongc@uw.edu; haoya@uw.edu, ckopca@uw.edu, yinhai@uw.edu); $^{2}$Z. Cui is in the School of Transportation Science and Engineering at Beihang University, China(e-mail: zhiyongc@buaa.edu.cn); $^{3}$S. Tien is a summer intern student at STAR Lab(e-mail: sophiejytien@hotmail.com)}%
}

\begin{document}

\setlength{\abovedisplayskip}{7pt}
\setlength{\belowdisplayskip}{7pt}

\maketitle
\thispagestyle{empty}
\pagestyle{empty}

\begin{abstract}
With accurate and timely traffic forecasting, adverse traffic conditions can be proactively predicted to guide agencies and network users to respond appropriately. However, existing work on traffic forecasting has primarily relied on historical traffic patterns, confining to short-term prediction (e.g., under 1 hour). To better manage future roadway capacity and accommodate social and human impacts, proposing a flexible and comprehensive framework to predict physical-aware, long-term traffic conditions for network users and transportation agencies is crucial. In this paper, the gap in robust long-term traffic forecasting was bridged by including social media features. A correlation study and a linear regression model were implemented to evaluate the significance of the correlation between two time-series data sets, (1) traffic intensity and (2) Twitter data intensity. These data sets were then fed into a proposed social-aware framework, named the Traffic-Twitter Transformer, which integrated Nature Language representations into time-series records for long-term traffic prediction. Experimental results in the Great Seattle Area showed that the proposed model outperformed baseline models in all evaluation matrices. This Nature Language Processing-joined social-aware framework promises to become a valuable tool of network-wide traffic prediction and management for traffic agencies. 
\end{abstract}


\section{INTRODUCTION}

\begin{figure}
  \begin{subfigure}[b]{0.12
  \textwidth}
    \centering
    \captionsetup{justification=centering}
    \includegraphics[width=\textwidth]{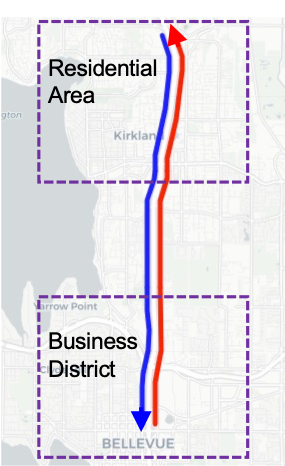}  
    \caption{Different land use locations}
    \label{fig:different_location}
  \end{subfigure}
  \begin{subfigure}[b]{0.35\textwidth}
    \centering
    \captionsetup{justification=centering}
    \includegraphics[width=\textwidth]{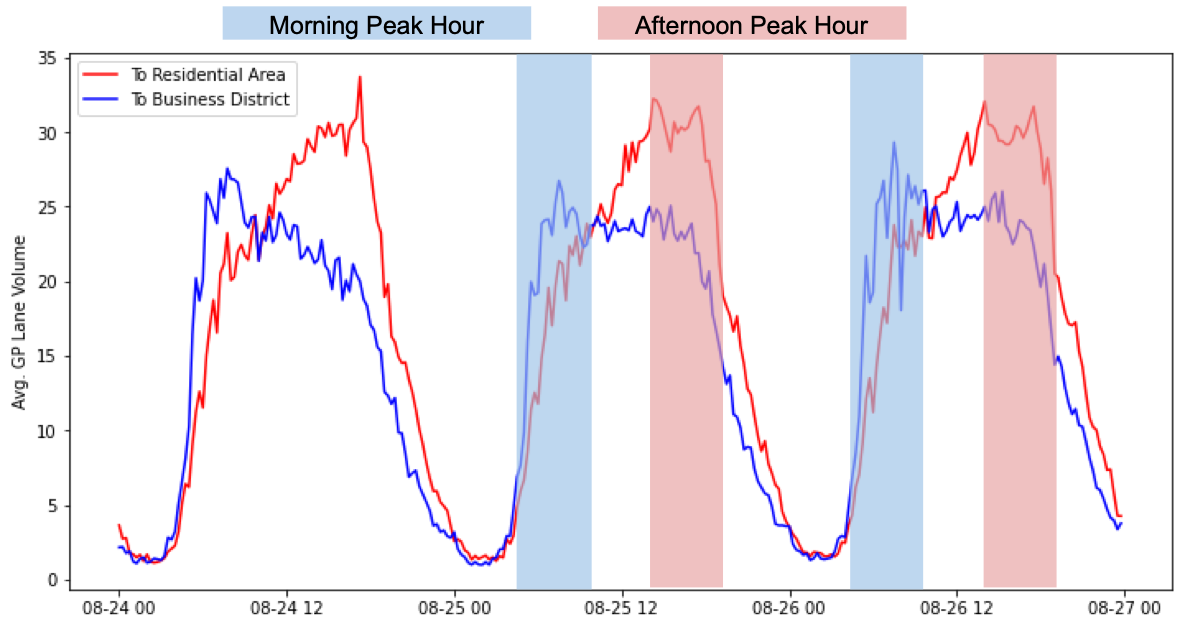}
    \caption{Traffic pattern variations under different land-use scenarios over morning/evening-peak hours}
    \label{fig:morning_afternoon}
  \end{subfigure}
  \caption{Geographical attributes and mutual relationship}
  \label{fig:spatial-temporal}
  \vspace{-1.7em}
\end{figure}

Nowadays, user demand (for both people and goods) has increased significantly. According to a report from Deloitte \cite{2017industryreport}, urban freight delivery will surge by more than 40\% by 2050. Cities are growing, and with this growth, are becoming more congestion; driving the need for intelligent solutions to address forthcoming traffic-related challenges. Intelligent Transportation System (ITS) offers data-driven approaches to improve traffic performance and efficiency. Reliable and accurate traffic forecasting, one facet of ITS \cite{boukerche2020artificial}, is a tool that has the potential to increase roadway capacity and alleviate congestion \cite{cheng2020mitigating}. Traffic forecasting is a spatial-temporal sequence prediction problem as shown in Fig. \ref{fig:spatial-temporal}, which involves both location adjacency and temporal impact. With regard to geographic attributes and social activities (Fig. \ref{fig:different_location}), traffic patterns might vary significantly (Fig. \ref{fig:morning_afternoon}). Multiple algorithms have taken these key points into consideration for predicting traffic parameters, including speed and volume \cite{vlahogianni2014short, liao2018deep, wu2019graph, cui2019traffic, zheng2020gman, guo2021learning}. These algorithms have made notable advancements in various aspects of traffic forecasting, including improving model accuracy and reducing computation time.

Through advancements in methods and increased accessibility of traffic data, deep learning models have outperformed conventional methods in traffic forecasting tasks \cite{ cai2016spatiotemporal, chikkakrishna2019short}. Several of these new methods include Recurrent Neural Networks (RNN), Long Short-Term Memory (LSTM), Convolutional Neural Networks (CNN), and Graph Neural Network (GNN). RNNs have received significant attention in handling sequential forecasting problems. However, with a deep network structure, the gradients being back-propagated have to go through consequent matrix multiplications based on the chain rule. In this case, RNN models learns to recognize sequential patterns, but the gradient tends to either blow up or vanish \cite{bengio1994learning}. LSTM models have the capability to address these gradient vanishing and explosion issues \cite{hochreiter1997long}. Thus, many studies in the last decade have leveraged LSTM \cite{tian2015predicting, vinayakumar2017applying, huang2017study, madan2018predicting, xiangxue2019data, tsai2021multivariate} to maintain a time-related sequence and internal memory with the loop structure. CNNs, which are powerful image processing algorithms that can effectively extract informative features, have been generalized to capture spatial relationships in traffic networks. Works from Ma et al. \cite{ma2017learning}, Zhang et al. \cite{zhang2019gcgan}, and Huang et al. \cite{huang2020deep}, each used CNN to detected traffic patterns from geographic figures as well as traffic time-space speed matrix and further forecast future speed. Liu et al. \cite{liu2017short}, Bogaerts et al. \cite{bogaerts2020graph}, and Ma et al. \cite{ma2020forecasting}, further combined stacked CNNs to extract spatial features with LSTM to integrate temporal information of traffic data. However, standard CNN-based approaches are incapable of dealing with various topological structures of traffic networks. To address this issue, researchers began to train traffic networks as a graph and applied GNNs to extract patterns from graph-structure data \cite{li2017diffusion, cui2019traffic}. Most of the aforementioned methods focused on short-term forecasting, with time periods ranging from 5 minutes to 1 hour ahead. Moreover, unanticipated complications, such as accidents and the COVID-19 pandemic, significantly affect prediction results. In this case, other meaningful auxiliary data, such as social media data, were examined to fill the gap for robust long-term forecasting.


Social media has evolved considerably in the past decade and is now extensively used to share user-generated information, ideas, sentiments, and other forms of expression \cite{campos2015tweets}. Thus, Twitter has become a powerful tool for gathering information from a reasonably large and diverse user pool. Since tweets can be retrieved in real-time with a relatively small building and maintenance costs, this data source can be treated as another type of sensor, such as loop detectors, for traffic conditions. A Seattle Mariners game, for example, may impact traffic near the T-mobile park (the stadium where the Mariners play). Those attending the game may post tweets about proximate traffic, which indirectly contributes information that has the potential to help with traffic forecasting.

Various studies have attempted to integrate social media data into transportation research. He et al.\cite{he2013improving} examined the possibility of using rich information in online social media to improve traffic prediction. The authors analyzed the correlation between traffic volume and tweet counts with various granularities. An optimization framework was also proposed to extract traffic indicators based on transformed tweet semantics. A recent work \cite{yao2021twitter}, dedicated to traffic forecasting with multi-source data features, considered leveraged machine learning models with tweet semantics to predict morning traffic conditions. The experimental results showed the capability of robustly learning traffic patterns from tweets semantics when compared to other algorithms with the proposed approaches. Other studies \cite{zhang2018deep, gu2016twitter} attempted to examine the relationships between words in tweets and traffic accidents. They identified important key terms in tweets to determine traffic incidents through tweet data mining.

Based to the previously discussed research, some inherent limitations are summarized, which provide the motivation to make innovations:
\begin{itemize} 
    \item \textbf{Ignore the culture impacts in prediction tasks.} Most of the previous research took common features, such as speed, volume, weather conditions, and roadway geometry, into consideration for traffic forecasting. Yet, cultural-related events (e.g., Black Lives Matter's rallies), which can severely influence traffic conditions, should also be considered.
    \item \textbf{Social impacts are not considered in the predicting values.} In some of the previous research, social information is considered as part of the input. However, the final output still mirrors classical traffic patterns (i.e., volume and speed), with less consideration of physical attributes combination. New traffic attribute matrices need to be proposed and further integrated with human activities.
    \item \textbf{Lack of a flexible long-term traffic forecasting approach.} Prior studies mainly focused on short-term forecasting. It is hard to model long-term spatial and temporal trends even using graph-based algorithms. The graph-structure connectivity also constrains the flexibility of data aggregation for each road segment. To tackle this problem, a new, flexible comprehensive model, that incorporates temporal-spatial dependence and tweet information, is required.
\end{itemize}

Inspired by the previously discussed research and their limitations, this study expands the temporal scale by predicting both morning-peak traffic \cite{yao2021twitter} and the whole day network-wide traffic performance. This work proposes a Nature Language Processing (NLP)-joined social-aware traffic forecasting model, originated from Transformer \cite{vaswani2017attention}, with a temporal encoder as opposed to a positional encoder, and includes social media features. Tweet data and traffic data can each be fed into the proposed model to predict an accurate long-term network-wide traffic performance. Unlike previous studies, this work utilizes a more interpretable and inclusive matrix, called the Traffic Performance Score (TPS) \cite{cui2020traffic}, to evaluate traffic conditions. Due to the integration of social media features, which contain personal opinions, the TPS (which ranges from 0\% to 100\%), is a surperior explanatory and comprehensive matrix to assess network-wide traffic states, rather than predicting classical metrics, such speed or volume.

In summary, the contributions of this paper are listed as follows:

\begin{itemize}
    \item A correlation study is conducted to prove that it is meaningful to involve social media features in the model, with a strong correlation between traffic data and Twitter data.
    \item An NLP-joined social-aware framework, Traffic-Twitter Transformer is proposed to increase robustness under various unexpected events.
    \item The forecasting results rely on historical traffic patterns as well as varied social media features to improve the model's flexibility.
    \item A time encoder is applied to replace the positional encoder in the originally developed Transformer. The time encoder allows the model to keep the recurrent characteristic from the sequential time-series input.
    \item An ablation study shows the causality of the proposed model and how each Twitter feature impacts traffic forecasting.
\end{itemize}

\section{Literature Review}

Time series analysis has been frequently used in traffic forecasting, which is essential for the success of ITS because of the periodicity of traffic conditions. However, only limited-scaled data sets may be utilized in traditional statistical models. Li et al. \cite{zhi2008improved} and Chan et al. \cite{chan2011neural} proposed an exponential smoothing model to predict traffic speed. Williams and Hoel \cite{williams2003modeling}, Han et al. \cite{han2004real}, and Kumar and Vanajakshi \cite{kumar2015short}, implemented Autoregressive Integrated Moving Average (ARIMA). Chikkakrishna et al. \cite{chikkakrishna2019short} employed ARIMA fitted with seasonal components (SARIMA) and Facebook Prophet, which was proposed by Taylor and Letham \cite{taylor2018forecasting}, to make predictions. Sun et al. \cite{sun2003use} stated that their proposed local linear regression has a better performance in comparison to k-nearest neighbors (KNN) and kernel smoothing estimator. The two approaches have been ameliorated by Cai et al. \cite{cai2016spatiotemporal} and Haworth et al. \cite{haworth2014local} respectively. Nonetheless, most of the aforementioned models do not properly employ many-to-many predictions. That is, they are not suitable for network-wide traffic state forecasting because these methods cannot process high-dimensional features and model complex spatial-temporal dependency. Therefore, many researchers have moved their attention to Artificial Neural Networks(ANN)-based approaches for achieving more desired results.

Recently, substantial advanced ANN-based models have been utilized in traffic forecasting. Since the traffic data are sequence-dependent, a variant of ANN was designed called RNN to tackle complications with time-series data. A simple recurrent network suffers from a fundamental problem of not capturing long-term dependencies in a sequence due to the vanishing gradient problem during the backpropagation process \cite{bengio1994learning}. LSTM was proposed to solve traffic prediction tasks while addressing this issue and showed outstanding performance \cite{hochreiter1997long}. Since the capability of this architecture can handle the problem of recurrent pattern, several studies modified and enhanced the original LSTM to become more robust \cite{zhao2017lstm, luo2019spatiotemporal, ma2015long}. For instance, a stacked bidirectional and unidirectional LSTM (SBULSTM) network \cite{cui2020stacked} was proposed to capture the forward and backward temporal dependencies in traffic data. LS-LSTM was developed to integrate both the latest trends and extra historical data patterns to achieve better performance \cite{tsai2021multivariate}. Moreover, some researchers also considered graph-related approaches in traffic forecasting to capture spatial-temporal dependency.  

Traffic networks can be identified as graphs with the combination of nodes and edges. Average speed and volume, for example, can be derived to depict the traffic conditions at each node, which each represents a road segment. Adjacency matrices, constructed based on road network connectivity, reveal the relationship between segments. Based on the pre-defined structure, Graph Convolutional Network (GCN)-based algorithms have been widely developed in traffic forecasting tasks recently \cite{yu2017spatio, yu2017spatiotemporal, zhang2019gcgan, cui2019traffic}. These works utilized a variety of extended GCNs to capture spatial inter-dependencies to improve computational efficiency \cite{wu2019graph} and enhance prediction performance \cite{cui2019traffic}. Recent efforts, such as GMAN \cite{zheng2020gman} and ASTGNN \cite{guo2021learning}, applied a more complex attention mechanism to capture dynamic spatial-temporal dependency. Although these novel algorithms achieved promising results, several complicates were left unresolved. For example, most GCN-based approaches can be recognized as a two-phase process. The first phase aggregates neighborhood awareness by GCN and the second applies transformed embeddings to prediction work. Through aggregating neighborhood knowledge, GNN may blur node/edge information, resulting in noisy representations of downstream prediction tasks \cite{leng2021enhance}. Furthermore, most existing traffic forecasting studies focus on short-term forecasting as Table \ref{short-term forecasting table} summarizes. The predicted future time frames are limited to 1 hour. To overcome the gap in long-term traffic forecasting, a model that can capture long-term sequence patterns, such as Transformer, should be investigated.

\begin{table}[h]
\caption{Short-term traffic forecasting works}
\vspace{-1.3em}
\label{short-term forecasting table}
\begin{center}
\renewcommand{\arraystretch}{1.5}
\begin{tabular}{c c}
\hline
\specialrule{.2em}{.1em}{.1em}
Predicted Timeframe & References\\
\specialrule{.2em}{.1em}{.1em}
\hline
less than 30 mins & \cite{williams2003modeling}, \cite{sun2003use}, \cite{han2004real}, \cite{zhi2008improved}, \cite{ma2015long}, \cite{cui2018deep}, \cite{luo2019spatiotemporal}, \cite{cui2020stacked}\\
\hline
30 mins - 1 hr	&	\cite{li2017diffusion}, \cite{yu2017spatio}, \cite{yu2017spatiotemporal}, \cite{zhao2017lstm}, \cite{wu2019graph}, \cite{guo2019deep}, \cite{zheng2020gman}, \cite{huang2020lsgcn},  
\cite{zheng2020hybrid}, \cite{guo2021learning}	\\
\specialrule{.2em}{.1em}{.1em}
\hline
\end{tabular}
\end{center}
\vspace{-1.3em}
\end{table}

Because AI methods have evolved significantly in Neural Language Processing (NLP), the Transformer \cite{vaswani2017attention}, a model with a multi-head attention mechanism, is proposed to help memorize long source sentences in neural machine translation and achieved maximal performance. In other words, the Transformer can solve forecasting tasks with a long sequence of time-series data. Broadly, the Transformer model can act as a GNN model that treats the entire sequence of input data as the local neighborhood, aggregating spatial-temporal correlations. Cai et al. \cite{cai2020traffic} developed a Traffic Transformer to leverage long-term temporal dependencies. Yan et al. \cite{yan2021learning} further enhanced the performance by learning the dynamic and hierarchical structure of traffic flow. However, due to the COVID-19 pandemic and several other unexpected culture-related events, the existing traffic prediction models are affected without consideration of related semantic information. Therefore, other meaningful data, such as social media data, need to be considered to bridge the gap of robust long-term traffic forecasting.

In this paper, an NLP-joined social-aware framework, Traffic-Twitter Transformer, is proposed, to predict traffic parameters under various unanticipated conditions. The forecasting results rely on historical traffic patterns and varied social media features to improve the model's robustness and flexibility.

\section{Data Preprocessing}

\subsection{Data Collection}

A combination of the Traffic Performance Score (TPS) dataset and the tweets dataset from May 1st, 2020 to August 31st, 2020, are analyzed.

\subsubsection{\textbf{TPS Dataset}}

The TPS dataset \cite{cui2020traffic} is collected from roughly 8000 inductive loop detectors deployed on the freeway network in the northwestern region of Washington State. The freeway network mainly includes several major freeways, such as I-5, I-90, I-405, and SR-520. It assigns the freeway to 106 segments and measures the average volume (Q) and average speed (V) of each segment with 15-minute intervals for calculating TPS. Traffic Performance Score (TPS), the prediction target of this research, is an indicator to measure the traffic performance of the traffic network. It is a value ranging from 0\% to 100\%. Overall network-wide traffic condition is the best when the TPS is 100\% and worsens when TPS closes to 0\%. 

\subsubsection{\textbf{Tweets Dataset}}

\begin{figure}[h]
    \centering
    \includegraphics[width=0.6\columnwidth]{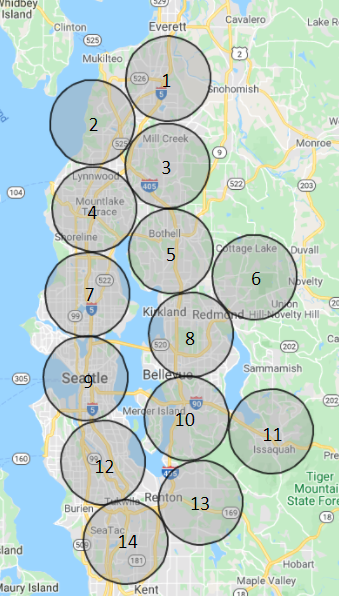} 
    \caption{Location segments of collected tweets}.
    \label{fig:tweet segment}
    \vspace{-1.7em}
\end{figure}

The tweets dataset is collected based on the location of the traffic network from the TPS dataset and Seattle's population distribution. Specifically, tweets from 14 sites are acquired, using a 5 kilometers buffer around the center of each segment from the TPS dataset, as shown in Fig. \ref{fig:tweet segment}. These tweets are then assigned to their respective segments, which are then applied to construct tweet feature matrices. The time interval is aligned with the 15-minute interval of the TPS dataset. Because tweets are clustered together, aggregation of the features from each tweet is required.

\subsection{Tweet Text Processing}

Three semantic features are extracted from the Tweet dataset: a term frequency feature, an accident-related feature, and a culture-related feature. 

\begin{figure}[h]
    \centering
    \captionsetup{justification=centering}
    \includegraphics[width=0.9\columnwidth]{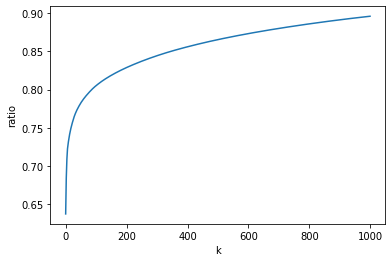} 
    \caption{Explained variance ratio from k = 1 to 1000 with \texttt{TruncatedSVD}}.
    \label{fig:explained variance}
    \vspace{-1.7em}
\end{figure}

Firstly, term frequency reflects specific traffic conditions. Specific terms, such as "congestion" and "roadblock," may correspond to traffic states, which is valuable information for traffic forecasting. In order to convert tweet terms to a numerical matrix, we transformed them into a high-dimensional matrix of token counts. The document-term frequency matrix shows that the vocabulary size was 91812 (91812 unique tokens.) When a threshold of three was established to filter out those words with low counting frequency, the dimensions of the frequency matrix were too large and computationally burdensome. Therefore, truncated singular value decomposition (SVD), a dimension reduction approach, was applied for our following computations. Truncated SVD is similar to traditional SVD methods, yet works well on sparse matrices like count and TF-IDF matrices \cite{halko2010finding}. The desired dimension for the frequency matrix was identified as $k = 100$, which explains approximately 80\% of the variation as shown in Fig. \ref{fig:explained variance}. To indicate the final term frequency for a segment at a certain time, all the transformed document-term frequencies were simply summed as a single value.

Second, traffic accidents can directly affect transportation operations by breaking down the segment-wise traffic. Therefore, accident-related features require extraction to better represent traffic conditions. Specifically, the number of accident-related tweets was counted based on accident-related keywords, listed in Table \ref{accident-related table}, as proposed by Zhang et al.\cite{zhang2018deep}.

\begin{table}[h]
\caption{Accident-related keywords}
\vspace{-1.3em}
\label{accident-related table}
\begin{center}
\begin{tabular}{p{8cm}}
\hline
\specialrule{.2em}{.1em}{.1em}
police, accident, traffic, crash, road, car, vehicle, highway, driver, county, injured, injuries, scene, hospital, died, patrol, morning, happened, dead, driving, department, involved, vehicles, passenger, hit, truck, monday, left, lane, killed, struck, closed, investigation
\\ \specialrule{.2em}{.1em}{.1em}
\hline
\end{tabular}
\end{center}
\vspace{-1.3em}
\end{table}

\begin{figure}[h]
  \centering
  \includegraphics[scale = 0.37]{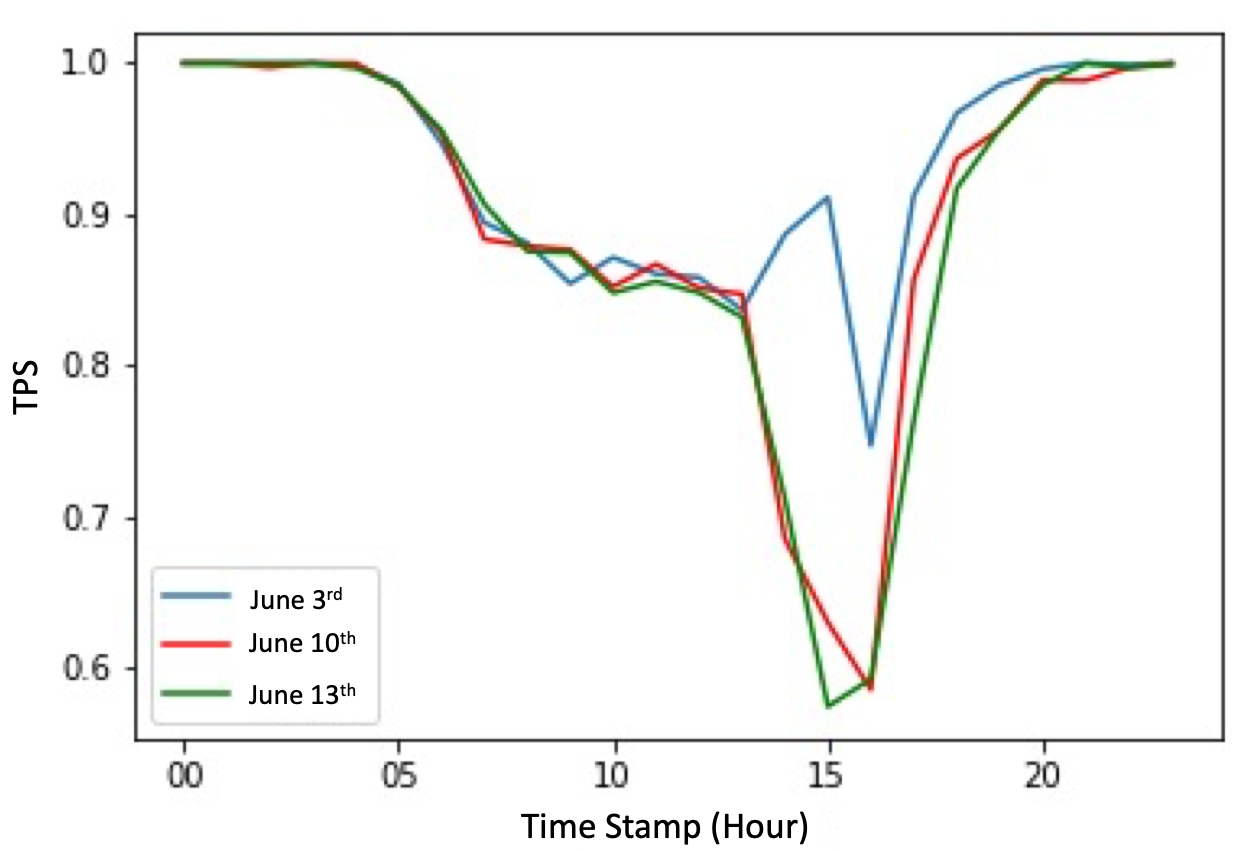}
  \caption{Difference of TPS at Downtown Seattle in 3 weeks}
  \label{fig:tps june}
  \vspace{-1.7em}
\end{figure}

Third, culture-related features were extracted as, they too, have the potential to greatly alter traffic conditions. For example, Fig. \ref{fig:tps june} demonstrates that June 3rd, 2020 (Wednesday) had a significantly different pattern than the same day of the two subsequent weeks (June 10th, 2020 and June 17th, 2020) in Downtown Seattle. Certain high-frequency words were discovered from twitter data, as shown in Table \ref{culture-related table}. These words are closely tied to \textit{Black Live Matters} and \textit{Defund Seattle Police} rallies held in the Downtown and Capitol Hill neighborhoods of Seattle on June 3rd, 2020. As a result, there were additional, and unanticipated, road traffic controls implemented in and around these neighborhoods. Culture-related features in this study were derived by counting the number of tweets containing culture-related keywords.

\begin{table}[h]
\caption{Culture-related keywords}
\vspace{-1.3em}
\label{culture-related table}
\begin{center}
\begin{tabular}{p{8cm}}
\hline
\specialrule{.2em}{.1em}{.1em}
blm, BlackLivesMatter, Ahmaud Arbery, Breonna Taylor, George Floyd, Jacob Blake, AllLivesMatter, protest, privilege, police, Seattlepd, Durkan, durkanresign, Anderson, mayorjenny, realdonaldtrump, seattlepd, hard, capitol, privilege
\\ \specialrule{.2em}{.1em}{.1em}
\hline
\end{tabular}
\end{center}
\vspace{-1.3em}
\end{table}

\subsection{Correlation Study}

In order to confidently include tweet features as inputs to predict traffic states, it is essential to confirm whether there is a statistically significant correlation between tweets and traffic features. Before the correlation analysis, it is practical to detrend time-series data with evident periodic fluctuations.

\subsubsection{\textbf{Data Detrending}}

\begin{figure}[h]
  \begin{subfigure}[b]{0.23\textwidth}
    \centering
    \includegraphics[width=\textwidth]{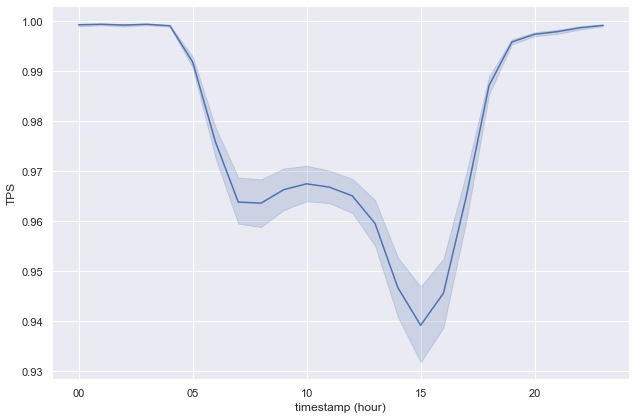}  
    \caption{Hourly traffic intensity}
    \label{fig:hourtps}
  \end{subfigure}
  \begin{subfigure}[b]{0.23\textwidth}
    \centering
    \includegraphics[width=\textwidth]{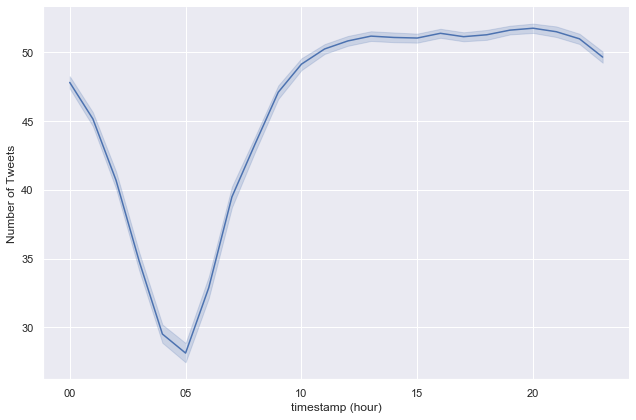}
    \caption{Hourly tweets intensity}
    \label{fig:hourtweet}
  \end{subfigure}
  
  \begin{subfigure}[b]{0.23\textwidth}
    \centering
    \captionsetup{justification=centering}
    \includegraphics[width=\textwidth]{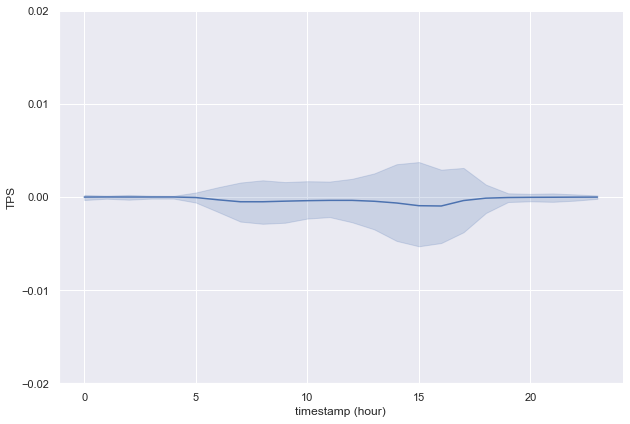}  
    \caption{De-trended hourly \\traffic intensity}
    \label{fig:detrendtps}
  \end{subfigure}
  \begin{subfigure}[b]{0.23\textwidth}
    \centering
    \captionsetup{justification=centering}
    \includegraphics[width=\textwidth]{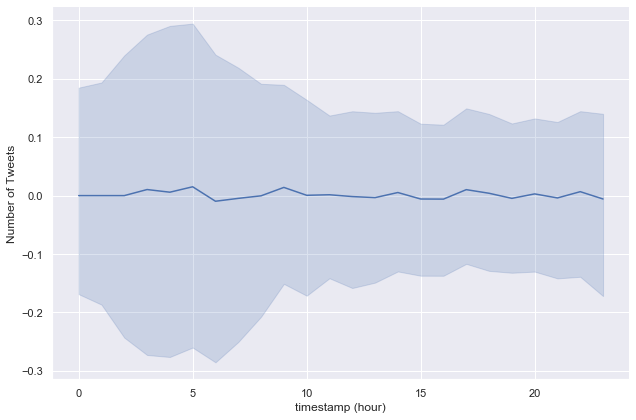}  
    \caption{De-trended hourly \\tweets intensity}
    \label{fig:detrendtweet}
  \end{subfigure}
  \caption{Traffic and Twitter data de-trending process visualization}
  \label{fig:detrend}
  \vspace{-0.8em}
\end{figure}

The periodic characteristic of traffic and social media pattern can be shown in Fig. \ref{fig:hourtps} and Fig. \ref{fig:hourtweet}. These seasonal fluctuations can strongly affect correlation studies and need to be removed. To detrend these patterns in time-series data, the process outlined in \cite{he2013improving} is followed. To exclude these patterns in the subsequent correlation analysis, the variation components were estimated and then subtracted from both traffic and tweet data to get the detrended version. with regard to the traffic data, the variation component $s_{h,d}$ can be formulated as:

\begin{equation}
    s_{h,d} = \frac{\sum_{\{t|(h,d)\}} \mathbf{v}^t}{|\{{t|(h, d)}\}|}
\end{equation}
\\
where $\mathbf{v} \in \mathbb{R}^T$ denotes the TPS data, T is the total number of time stamps, and its $t^{th}$ element $\mathbf{v}^t$ is the TPS averaged over all detectors in time stamp t. Besides, $|\{\cdot\}|$ denotes the number of elements in the set and pair (h, d) consists of $h = 0, . . . , 23$ and $d = 0, . . . , 6$.

Thus, the detrend TPS $\mathbf{v}^\prime \in \mathbb{R}^T$ can be defined as followed:
\begin{equation}
    \mathbf{v}^\prime = \mathbf{v}^t - s_{h, d}.
    \label{detrend equation}
\end{equation}
The Twitter data $\mathbf{c} \in \mathbb{R}^T$ follows the same procedure to access the detrended version $\mathbf{c}^\prime$.

Fig. \ref{fig:detrend} shows the comparison of the original data and the detrended data for both traffic and tweet intensities. The recurrent patterns in Fig. \ref{fig:hourtps} and Fig. \ref{fig:hourtweet} can be clearly observed. After detrending by Equation \ref{detrend equation}, both their original periodic patterns, as showed in Fig. \ref{fig:detrendtps} and Fig. \ref{fig:detrendtweet}.

\subsubsection{\textbf{Correlation Analysis}}

\begin{figure}
  \begin{subfigure}[b]{0.5\textwidth}
    \captionsetup{justification=centering}
    \includegraphics[width=\textwidth]{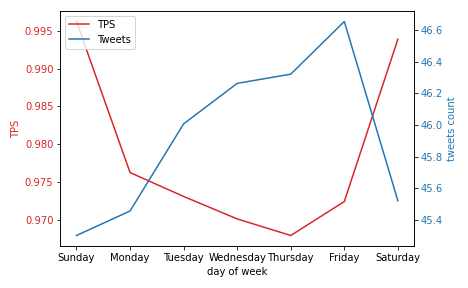}  
    \caption{Traffic and tweets intensity by day of week}
    \label{fig:dayofweek}
  \end{subfigure}
  \begin{subfigure}[b]{0.5\textwidth}
    \captionsetup{justification=centering}
    \includegraphics[width=\textwidth]{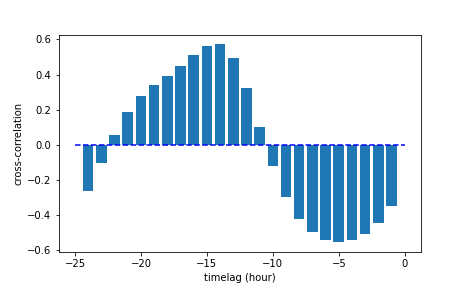}
    \caption{Cross correlation between traffic tweets intensity}
    \label{fig:cross}
  \end{subfigure}
  \caption{Correlation analysis}
  \label{correlation}
\end{figure}

As a first step towards predicting traffic intensity using Twitter data, the correlation between the social activity and the traffic intensity measure is investigated. The relationship between average TPS and average tweets counts across the days of the week is illustrated in Fig. \ref{fig:dayofweek}. The Pearson correlation between TPS and tweet counts is -0.223, indicating a negative relationship between the two variables, with TPS increasing as tweet count drops.

The time lag effect of TPS and tweet counts were also considered in the correlation analysis. Fig. \ref{fig:cross} shows the cross-correlation results between the current detrended traffic intensity $\mathbf{v}^\prime$ and the detrended tweets activity intensity $\mathbf{c}^\prime$ over the previous 24 hours with a time resolutions of 1 hour. The height of the blue bar at time lag $− \delta t$ represents the correlation between $\mathbf{v}^{\prime t}$ and $\mathbf{c}^{\prime t-\delta t}$. The cross correlation is negative when the time lag within 10 hours, and shows a sine wave over the time. This implies that when twitter activity is lower than usual, traffic performance on the road network is higher (better) than the average in the near future, assuming a 10-hour time lag.

Further, the significance of the correlation between the two time-series data sets are evaluated by combining historical traffic intensity ($\mathbf{v}^{\prime t-\delta t}$), and current and lagged tweet intensity ($\mathbf{c}^{\prime}$ and $\mathbf{c}^{\prime t-\delta t}$), to the auto-regressive model for current traffic intensity prediction \cite{smith1997model}. Specifically, current traffic intensity $\mathbf{v}^\prime$ can be predicted by the following linear regression model.

\begin{equation}
    \mathbf{v}^{\prime t} = \alpha + \beta_1 \mathbf{v}^{\prime t-1}  + \beta_2 \mathbf{c}^{\prime t} + \beta_3 \mathbf{c}^{\prime t-1} 
\end{equation}
where $\alpha$ is the intercept, $\beta_1, \beta_2, \beta_3$ are coefficients associated with traffic and Twitter data with various lags. The results are shown in Table \ref{linear regression table} with an $R^2$ as 0.776. The detrended tweets count with one-hour time lag ($\mathbf{c}^{\prime t-1}$) prove to be statistically significant with a p-value of 0.041. The negative correlation between TPS and tweets data is consistent with the previous analysis.

The results of correlation analysis in Table \ref{linear regression table} confirm a statistically significant relationship between TPS and tweets characteristics. As a result, it is meaningful to incorporate twitter features into the proposed model for network-wide traffic forecasting.

\begin{table}[h]
\caption{Results of Linear Regression for correlation study}
\vspace{-1.3em}
\label{linear regression table}
\begin{center}
\renewcommand{\arraystretch}{1.5}
\begin{tabular}{c cccc}
\hline
\specialrule{.2em}{.1em}{.1em}
Coefficient & Value & Std. Error & p-value\\
\specialrule{.2em}{.1em}{.1em}
\hline
$\alpha$	&	-0.0013 & 0.000 & 0.90	\\
\hline
$\beta_1$	&	0.8809 & 0.009 & 0.00*	\\
\hline
$\beta_2$	&	-0.0640 & 0.004 & 0.095	\\
\hline
$\beta_3$	&	-0.0844 & 0.041 & 0.041*	\\
\specialrule{.2em}{.1em}{.1em}
\hline
\end{tabular}
\end{center}
\vspace{-1.3em}
\end{table}

\section{Model Preliminaries} \label{Preliminaries}

Changes in traffic conditions for any reason, can propagated (shockwave) congestion backward and forward and potentially impact connecting road segments \cite{nguyen2016discovering}. Thus, to consider robust traffic forecasting, network-wide traffic data matrices must be utilized \cite{cui2020stacked}. For this work, a traffic data matrix is constructed through historical data collected from segment sensors across the entire network. The matrix consists of $N$ previous timestamps before time $T$ in $M$ segments. Each element $x_{t}^m$ symbolizes the traffic condition in $m^{th}$  segment at $t^{th}$ timestamp:

\begin{equation}
    X^{T}_M = 
    \begin{bmatrix}
    x^{T -N}_{1} & x^{T -N}_{2} & ...  & x^{T -N}_{M} \\
    x^{T -N-1}_{1} & x^{T -N-1}_{2} & ...  & x^{T -N-1}_{M} \\
    \vdots & \vdots & \ddots & \vdots\\
    x^{T -1}_{1} & x^{T -1}_{2} & ...  & x^{T -1}_{M} \\
    \end{bmatrix}
    \label{eq:def_of_X}
\end{equation}

\begin{figure*}[b]
    \centering
    \includegraphics[scale = 0.38]{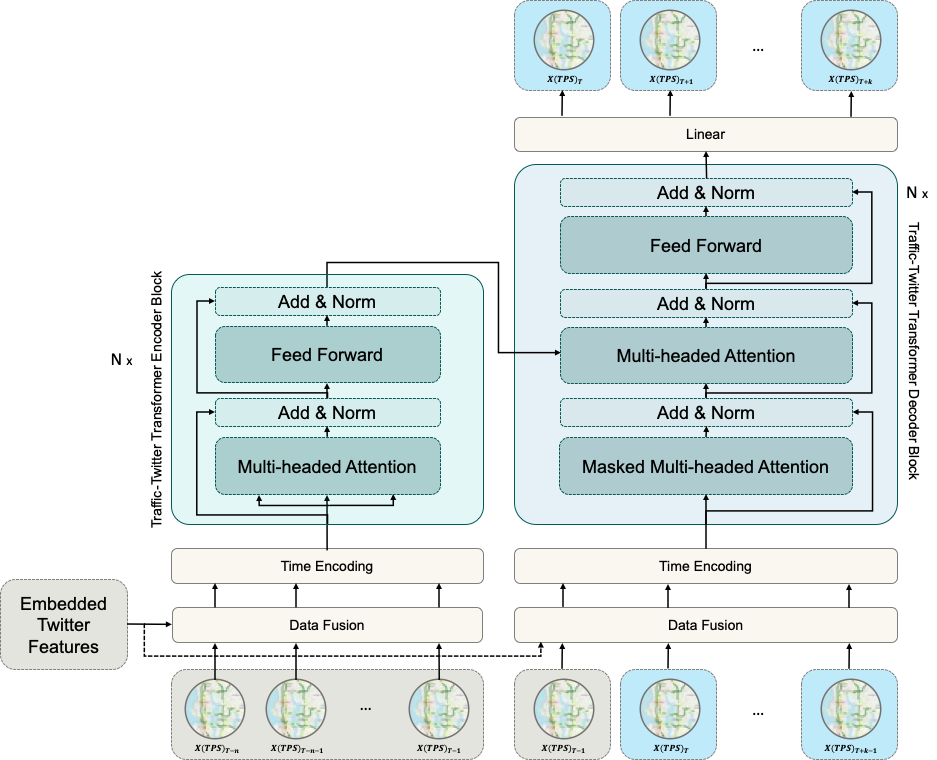}
    \caption{Traffic-Twitter Transformer architecture}
    \label{fig:archi}
    \vspace{-1.3em}
\end{figure*}

 Meanwhile, the twitter feature matrix is constructed using tweets from May 1, 2020, to August 31, 2020. First, tweets are collected and assigned to the nearest segment based on the distance between the center of the bounding circle and the segment. Then the remaining tweets are filtered for accident- and culture-related keywords. Ultimately, of the 687,803 original tweets, 150,736 accident-related tweets remained, and 46,276 culture-related tweets remained. The matrix $C \in \mathbb{R}^{N \times (K \times M)}$ can be defined as the following equation:
 
 \begin{equation}
    C^{T}_{K \times M} = 
    \begin{bmatrix}
    C_1 {}^{T}_{M} & C_2 {}^{T}_{M} & ...  & C_K {}^{T}_{M}\\
    \end{bmatrix}
    \label{eq:def_of_D}
\end{equation}
 where $C_1 {}^{T}_{M}, C_2 {}^{T}_{M}, ..., C_K {}^{T}_{M} \in \mathbb{R}^{N}_M$. Each of the submatrices represents one type of tweet count matrix consisting of $N$ previous timestamps before time $T$ in $M$ segments. More specifically, accident-related and culture-related tweets are collected in the Greater Seattle area; therefore, N should be 2 to describe two types of features that were included in this study.

The research problem can be described as: given the historical traffic data matrices and twitter feature matrices with time series $[T - N, T - N + 1, …, T - 1]$, cab the network-wide traffic condition $\hat{Y}^T_{M}$ in $M$ segments at timestamp $T$ be predicted? The procedure developed in this work to forecast network-wide traffic conditions can be presented as:\\

\begin{equation}
    \hat{Y}^T_{M} = G(X_{M}^T, C)
    \label{eq:def_of_hat_y}
\end{equation}
where $G$ is the traffic-twitter transformer, the primary algorithm proposed in this study. The objective is to minimize the difference between the sequence of predicted TPS $\hat{Y}^M_{T}$ and the ground truth data $Y^M_{T}$.

\section{Methodology} \label{methodology}

Inspired by a well-known Natural Language Processing (NLP) model, Transformer \cite{vaswani2017attention}, the original architecture has been modified for the task of long-term network-wide traffic forecasting. The proposed Traffic-Twitter Transformer model, as shown in Fig. \ref{fig:archi}, consists of three main blocks: Data fusion block, Traffic-Twitter Transformer encoder block, and decoder block:

First, traffic-related features and Twitter features are fused together. Here column-wise concatenation is used to transform them into a long vector $\tilde{X}_t$ in the data fusion block, as follows:

\begin{equation}
    \bar{X}_t = X_t \oplus C_1 \oplus C_2 \oplus C_3 
    \label{eq:def_of_bar_x_t}
\end{equation}
where $C_1$ is the tweet term frequency feature, $C_2$ is the accident-related tweet feature, and $C_3$ is the culture-related tweet feature. Each aggregated long vector represents multiple features in a specific timestamp, such as a TPS score, traffic volume, average speed, and corresponding Twitter features.

The sequence of fused data is then encoded with the time encoder. This series of data is entered into the Traffic-Twitter Transformer model concurrently, unlike the RNN-based model. The input elements in the original transformer model only notify the model about the input order depending on their index in the sequence. To account for the order of the traffic information in the input sequence and also maintain the temporal recurrent characteristic \cite{abnar2020transferring}, a time-encoded vector is added to each input data. These vectors are generated from the time encoder, which replaces the positional encoder in the original Transformer architecture.

The time encoder generator can be separated into two parts. The first requires generation of the original positional encoder features as formulated:

\begin{equation}
    \tau_t(k) = 
        \begin{cases}
            sin(pos/10000^{k/d_{\tau}}) & \text{, k is even}\\
            cos((pos/10000^{(k-1)/d_{\tau}}) & \text{, k is odd}\\
        \end{cases} 
    \label{eq:def_of_time_encoder}
\end{equation}
where $\tau_t(k)$ is the $k^{th}$ feature,  $pos$ is the position and $d_{\tau}$ is the dimension of the encoder features. For the 2D positional encoding matrix, the size of the row is the length of the input sequence, and the number of columns would be equal to the number of input features. For example, 12-steps of historical data are used with 64 different features to predict future traffic conditions. The shape of the positional encoding matrix would be $[12\times64]$.

Meanwhile, the second part of the time encoder requires extraction of seven-dimensional time features and combining them into the time encoder. The normalized seven-dimensional time features are \emph{minute}, \emph{hour}, \emph{dayofweek}, \emph{day}, \emph{dayofyear}, \emph{month}, and \emph{weekofyear}. Specifically, the input embedding sequence is concatenated with the positional encoder features and the additional time features as followed:

\begin{equation}
    \tilde{X}_t = W (\bar{X}_t \oplus \tau_t \oplus T_t)
    \label{eq:def_of_tilde_x}
\end{equation}
where $X_t$ is the input embedding sequence, $W \in \mathbb{R}^{d_{\tau} \times (2d_{\tau} + Dim_t)}$ is the weight matrix, $Dim_t$ is the temporal feature dimension, $T_t$ is the time feature, and $\oplus$ is the symbol of concatenation. Finally, the processed input sequences are fed into the encoder block to further extract inherent traffic states.


Then, the input vectors are sent to a Multi-headed Attention cell in the encoder block, which accesses all input sequences for hints that might aid in better encoding traffic semantics. Different learnable neural networks can project the encoded sequence of input into three matrices (Query (Q), Key (K), and Value (V)) in the attention mechanism:
\begin{gather}
    Q = \tilde{X}_t W_Q \\
    K = \tilde{X}_t W_K \\
    V = \tilde{X}_t W_V
    \label{eq:self_attention}
    \vspace{-10em}
\end{gather}
The output can be generated as a weighted sum of the values using the aforementioned matrices. The weight allocated to each value is determined by a compatibility function of the query with the relevant key: 
\begin{equation}
    Attention(Q, K, V) = softmax(\frac{QK^T}{\sqrt{d_k}})V
    \label{eq:self_attention_output}
\end{equation}
where $Q, K, V \in \mathbb{R}^{N \times d_k}$, and $d_k$ can be utilized as a scaling factor which provides a more stable gradient to the model \cite{vaswani2017attention}. It can be further expanded to multi-headed attention. By repeating the attention mechanism $h$ times, the multi-headed output can be attained as:
\begin{equation}
    \begin{aligned}
        MultiHeaded(Q, K, V) = Concat(head_1, ..., head_h)W^O \\
        head_i = Attention(QW_i^Q, KW_i^K, VW_i^V)
        \label{eq:self_attention_output_2}
    \end{aligned}
\end{equation}
where $d_k = d_v = d_{model} / h$, $W_i^Q, W_i^K, W_i^V \in \mathbb{R}^{d_{model} \times d_k}$, and $W^O \in \mathbb{R}^{hd_v \times d_{model}}$. In this study, the number of heads (h) is set as 8. 

After processing through a multi-headed attention cell, the processed data is passed to a position-wise fully connected feed-forward network. The representation data generated by the encoder block is sent to the decoder block as input data.

The decoder block takes the representation data received from the encoder and the last timestamp of traffic-related data $X_{T-1}$ as inputs to perform a step-wise prediction multiple steps ahead for future traffic conditions. A masked attention mechanism is applied to mask future positions to prevent data leakage and only allow attending the earlier positions in the output sequence. A linear feed-forward network is assigned as the last step to improve the expressiveness of the model and reshape the output to be the same as the sequence of label data. The outcome of each step is fed to the bottom decoder in the next step, and the decoder use the inputs to predict the next timestamp results.

After developing the aforementioned technical approaches, three widely-used metrics in traffic forecasting (Mean Squared Error (MSE), Mean Absolute Error (MAE), and Mean Absolute Percentage Error (MAPE))  were selected to evaluate the model accuracy, which can be calculated using the following equations:
\begin{gather}
    MSE = \frac{1}{n}\sum_{i=1}^n (y_i - \hat{y}_i)^2
    \label{eq:def_of_mse}\\
    MAE = \frac{1}{n}\sum_{i=1}^n |y_i - \hat{y}_i|
    \label{eq:def_of_mae}\\
    MAPE = \frac{1}{n}\sum_{i=1}^n \left| \frac{y_i - \hat{y}_i}{y_i}\right| \times 100\%
    \label{eq:def_of_mape}
\end{gather}
where $y_i$ and $\hat{y}_i$ represent the ground truth and predicted value respectively.

\section{Experiments}

Each data set was collected at a 15-minute interval in the Greater Seattle area from May 2020 to August 2020. Three months of data were used as the training dataset, and 15 days were used as the validation and the testing dataset respectively. In order to achieve the goal of long-term forecasting, methods suggested by Essien et al. \cite{essien2020deep} were used to set 12-steps ahead as a target for multi-step forecasting. Conclusively, an end-to-end long-term network-wide forecasting was executed, predicting the upcoming 12-steps ahead (3 hours ahead) given 12 historical sequences of data.

\subsection{Experiment Results Analysis and Comparison}

Based on the description of the model in Section \ref{methodology}, each input sequence  $X(TPS)_{i}$ is a 3D matrix with the shape $B\times1\times(Num\_of\_Segment \cdot 4)$ where B is the batch size and $Num\_of\_Segment$ is the total number of the segments in the network. It is a matrix with 12 historical sequences of data. The last dimension $Num\_of\_Segment \cdot 4$, consists of each segment's TPS, volume, average speed, and corresponding twitter features. The output data is also a 3D matrix with the shape $[B, 1, Num\_of\_Segment]$. Because the prediction model is many-to-many, the length of the output is 12. Therefore, the shape of the final concatenated output is $[B, 12, Num\_of\_Segment]$.

\subsubsection{\textbf{Comparison with Baseline Models for Network-wide Traffic Condition Prediction}}

Five baseline models that have the ability to predict the traffic condition of the whole network were selected to compare with the Traffic-Twitter Transformer model developed in this work: 

\begin{enumerate}
    \item a Stacked bidirectional and unidirectional LSTM (\textbf{S-LSTM}) with the default 2-layer LSTM structure setting \cite{cui2020stacked}; 
    \item an Attention graph convolutional sequence-to-sequence model (\textbf{AGC-Seq2Seq}), which incorporates GCN and RNN modules to capture the spatial-temporal dependency \cite{zhang2019multistep};
    \item an Adaptive Graph Convolutional Recurrent Network (\textbf{AGCRN}), which combines Node Adaptive Parameter Learning module and Data Adaptive Graph Generation module with recurrent networks for the multi-step traffic prediction task \cite{bai2020adaptive}; 
    \item a Vanilla Transformer (\textbf{Transformer}) first developed in NLP-based research then applied to various applications, including transportation engineering. With an attention mechanism, Transformer can be applied to longer time-series forecasting tasks \cite{vaswani2017attention};and
    \item a Graph multi-attention network (\textbf{GMAN}) employs an encoder-decoder architecture, in which both the encoder and the decoder are made up of numerous spatio-temporal attention blocks, to model the traffic states \cite{zheng2020gman}.
\end{enumerate} 

We applied the default settings for the above baseline models from their original studies. It is important to note that, without the data fusion block (Equation \ref{eq:def_of_bar_x_t}), the baseline models cannot combine traffic and twitter features. The proposed model, Traffic-Twitter, is the only model with the ability to benefit from this component in the following experiments.

\begin{table}[h]
    \caption{Network-wide Overall Performance Comparison of the Proposed Model with Baseline Models}
    \vspace{-1.3em}
    \label{overall result}
    \begin{center}
    \renewcommand{\arraystretch}{1.5}
    \begin{tabular}{c c c c}
    \hline
    \specialrule{.2em}{.1em}{.1em}
    Model & MSE & MAE & MAPE\\
    \specialrule{.2em}{.1em}{.1em}
    \hline
    S-LSTM	&	0.0028  & 0.0207  & 2.9168\% 	\\
    \hline
    AGC-Seq2Seq	&	0.0025 & 0.0202  & 2.7929\%	\\
    \hline
    AGCRN	&	0.0026 & 0.0201  & 2.7358\%	\\
    \hline
    Original Transformer	&	0.0021  & 	0.0160  &  2.3315\%  \\
    \hline
    GMAN	&	0.0021  & 	0.0151  &  2.2527\%  \\
    \hline
    \textbf{Our Model*}	&	\textbf{0.0019} & \textbf{0.0135} & \textbf{2.0141\%}	\\
    \specialrule{.2em}{.1em}{.1em}
    \hline
    \multicolumn{4}{r}{* Our Model: Traffic-Twitter Transformer}
    \end{tabular}
    \end{center}
    \vspace{-1em}
\end{table}

Table \ref{overall result} reflects the network-wide prediction from each of the six different models (the model proposed in this work and the five comparison models). The S-LSTM model can be considered the baseline for all deep learning-based approaches. Given a more complex architecture with an encoder-decoder structure, the AGC-Seq2Seq model performed better than the S-LSTM in all three metrics. The Transformer achieved a better performance than the AGC-Seq2Seq, and the GMAN model exhibits performance comparable to Transformer. This suggests the attention mechanism is capable of learning long time-series patterns and of achieving compelling results. The Traffic-Twitter Transformer presents the best score across all evaluation metrics, suggesting that it is more suitable for application in comparatively non-stationary traffic conditions.

The step-wise prediction results were further investigated to compare the performance in time steps ahead as follows: 15 minutes, 60 minutes, 120 minutes, and 180 minutes ahead. The results are shown in Table \ref{stepwise result}. The S-LSTM model displayed a competitive prediction result for the 15-minute ahead task; however, its prediction decreased significantly when additional timestep prediction processing was applied. This was anticipated because the S-LSTM can only access the latest former hidden state and cell state, which presents embedding information of previous steps. Thus, the prediction performance predictably decreases when processing a long-term prediction. The AGC-Seq2Seq model produced a similar finding but with slightly better outcomes. Its architecture, with a more reliable encoder-decoder structure, was likely the primary driver of the improvement.

Surprisingly, AGCRN outperformed all other models in a 15-minute prediction challenge regarding MSE performance. For long-term occupations, however, its forecast accuracy steadily decreased. In contrast, the Transformer model supplies all historical data concurrently and determines which timestamp data is crucial by giving those inputs  weights, thus generating a more competitive result. With a Transformer-like architecture that includes an encoder-decoder structure and a multi-attention mechanism, GMAN produced a comparable result. The model proposed in this work, the Traffic-Twitter Transformer, delivered the best performance across each of the four time steps. The proposed enhancements in the architecture, taking both traffic data and Twitter features into consideration, shows an effective result in complicated network-wide long-term traffic forecasting.

\begin{table*}[t]
    \caption{Step-wise Performance Comparison of the Proposed Model with Baseline Models}
    \vspace{-1.3em}
    \label{stepwise result}
    \centering
    \footnotesize
    \begin{center}
    \renewcommand{\arraystretch}{1.5}
    \begin{tabular}{c c c c c c c c c c c c c}
    \hline
    \specialrule{.2em}{.1em}{.1em}
    \multirow{2}{*}{Model} &  \multicolumn{3}{c}{15 min} &  \multicolumn{3}{c}{60 min} &  \multicolumn{3}{c}{120 min} &  \multicolumn{3}{c}{180 min} \\& MSE & MAE & MAPE & MSE & MAE & MAPE & MSE & MAE & MAPE & MSE & MAE & MAPE \\
    \specialrule{.2em}{.1em}{.1em}
    \hline
    S-LSTM & 0.0026 & 0.0195  & 2.0137\% & 0.0028 & 0.0225  &  2.2571\% &	0.0033 & 0.0232  &	2.5342\% &	0.0085 & 0.0238 & 5.0719\%  \\
    \hline
    AGC-Seq2Seq & 0.0023 & 0.0190 & 2.0212\% & 0.0025 & 0.0199 & 2.2250\% &	0.0026 & 0.0217 & 2.3297\% &	0.0032 & 0.0228 & 2.8438\% \\
    \hline
    AGCRN & \textbf{0.0012} & 0.0136 & 1.5375\% & 0.0023 & 0.0190 & 2.2901\% &	0.0029 & 0.0221 & 2.5123\% &	0.0033 & 0.0233 & 3.2968\% \\
    \hline
    Transformer & 0.0015 & 0.0138 &  1.6095\% & 0.0019 & 0.0143 & 1.6610\% &	0.0021 & 0.0169 &	1.8039\% &	0.0024 & 0.0175 & 2.0876\%	\\
    \hline
    GMAN & 0.0016 & 0.0141 &  1.7313\% & 0.0019 & 0.0158 & 1.8033\% &	0.0021 & 0.0159 &	1.8035\% &	0.0023 & 0.0164 & 2.0276\%	\\
    \hline
    \textbf{Our Model*} & 0.0013 & \textbf{0.0089} & \textbf{1.3878\%} & \textbf{0.0017} & \textbf{0.0105} & \textbf{1.4299\%} & \textbf{0.0018} & \textbf{0.0132} & \textbf{1.5645\%} & \textbf{0.0020} & \textbf{0.0136} & \textbf{1.7078\%} \\
    \specialrule{.2em}{.1em}{.1em}
    \hline
    \multicolumn{13}{r}{* Our Model: Traffic-Twitter Transformer}
    \end{tabular}
    \end{center}
    \vspace{-1.3em}
\end{table*}

\subsubsection{\textbf{Ablation Study of Twitter Features}}

In this experiment, each Twitter feature and time encoder are removed sequentially and the performance decrements are computed to determine how much each of them contributed to the model prediction. The results are shown in Table \ref{performance_decre_remove}.



The performance of the proposed model, the Traffic-Twitter Transformer, suffers the largest MSE decrement from 0.00186 to 0.00203 when the accident-related characteristics were disregarded. The removal of the time encoder also significantly impacts overall MSE performance. Contrarily, the MSE suffers relatively small decrement when either the frequency term or culture-related features is ignored. These results suggest the importance of inclusion of social media data, particularly tweets with accident-related semantics. They also validate the replacement of the original positional encoder with the time encoder. There are three important reasons for this: (1) The frequency-term features reflect a pattern of daily activity that combines all types of tasks, including current traffic states, none of which has a strong correlation to traffic conditions; (2) culture-related features have particular temporal and spatial characteristics, which have the potential to help improve particular cases, but less than the accident-related features, which mainly focus on traffic states in the network; and (3) the temporal features provide more crucial information in traffic forecasting tasks than positional indices.

\begin{table}[h]
    \caption{Ablation results of removing particular feature}
    \vspace{-1.3em}
    \label{performance_decre_remove}
    \begin{center}
    \renewcommand{\arraystretch}{1.5}
    \begin{tabular}{l c}
    \hline
    \specialrule{.2em}{.1em}{.1em}
    \multicolumn{1}{c}{Description}  & MSE\\
    \specialrule{.2em}{.1em}{.1em}
    \hline
    Remove culture-related feature	 & 0.00195	\\
    \hline
    Remove document-term frequency	&	 0.00199	\\
    \hline
    Remove accident-related feature	 & 0.00203	\\
    \hline
    Remove time encoder	 & 0.00204	\\
    \specialrule{.2em}{.1em}{.1em}
    \hline
    \multicolumn{2}{r}{* Proposed Traffic-Twitter Transformer MSE:  0.00186}
    \end{tabular}
    \end{center}
    \vspace{-1em}
\end{table}

\section{Conclusion and Future Work}

A novel NLP-joined social-aware transformer model, Traffic-Twitter Transformer, is proposed in this paper for network-wide traffic forecasting. The contributions concentrate on five aspects: (1) A traffic and Twitter data integrated structure is proposed to accommodate external interventions; (2) a real-world data set is used to evaluate the proposed method; (3) both spatial and temporal features are fused as input to improve the model robustness; (4) a time encoder is designed to replace the positional encoder to retain time dependency from data with strong temporal characteristics; and (5) Twitter features are summarized with great potential to help improve traffic condition prediction.

According to the correlation analysis, it is clear that Twitter features can be included as important factors that can impact traffic conditions, especially when unforeseen events happen. The experiment results reveal that Traffic-Twitter Transformer the most accurate model in predicting the overall network-wide traffic performance as compared to five competitive comparison models. The proposed model also delivers surperior result in all timestep-ahead predictions, which suggests the Traffic-Twitter Transformer can accommodate complicated spatial dependency and expand the ability to model long-term temporal dependence.

Future works can be divided into three aspects: (1) Twitter dataset can be further investigated to extract meaningful semantics, (e.g., local sports events); (2) more traffic network data sets can be used to evaluate the proposed model and validate its generalizability; and (3) self-supervised learning and other representation learning algorithms may be used to learn meaningful embeddings, which can help with a variety of downstream tasks, including traffic forecasting.

\bibliographystyle{ieeetr}
\bibliography{references.bib}

\end{document}